\documentclass[lettersize, journal]{IEEEtran}

\usepackage{graphics} 
\usepackage{epsfig} 
\usepackage{mathptmx} 
\usepackage{times} 
\usepackage{amsmath} 
\usepackage{amssymb}  
\usepackage{xcolor}
\usepackage{colortbl}
\usepackage{url}
\usepackage{gensymb}
\usepackage{algorithm}
\usepackage{algpseudocode}
\usepackage{listings}
\usepackage{caption}

\usepackage{hyperref}
\hypersetup{
    colorlinks=true,
    linkcolor=blue,
    filecolor=magenta,      
    urlcolor=blue
}
\def\BibTeX{{\rm B\kern-.05em{\sc i\kern-.025em b}\kern-.08em
    T\kern-.1667em\lower.7ex\hbox{E}\kern-.125emX}}
\usepackage{balance}

\newcommand{\commenting}[1]{}

\newcommand{\newAddition}[1]{{\color{magenta} NEW ADDITION: #1}}
\newcommand{\toShorten}[1]{{\color{blue} #1}}

\definecolor{Gray}{gray}{0.85}
\definecolor{LightCyan}{rgb}{0.88,1,1}
\definecolor{White}{rgb}{1,1,1}

\newcolumntype{g}{>{\columncolor{Gray}}c}

\colorlet{punct}{red!60!black}
\definecolor{background}{HTML}{EEEEEE}
\definecolor{delim}{RGB}{20,105,176}
\colorlet{numb}{magenta!60!black}

\lstdefinelanguage{json}{
    basicstyle=\normalsize\ttfamily,
    numberstyle=\scriptsize,
    stepnumber=1,
    numbersep=8pt,
    showstringspaces=false,
    breaklines=true,
    frame=lines,
    backgroundcolor=\color{background},
    literate=
     *{0}{{{\color{numb}0}}}{1}
      {1}{{{\color{numb}1}}}{1}
      {2}{{{\color{numb}2}}}{1}
      {3}{{{\color{numb}3}}}{1}
      {4}{{{\color{numb}4}}}{1}
      {5}{{{\color{numb}5}}}{1}
      {6}{{{\color{numb}6}}}{1}
      {7}{{{\color{numb}7}}}{1}
      {8}{{{\color{numb}8}}}{1}
      {9}{{{\color{numb}9}}}{1}
      {:}{{{\color{punct}{:}}}}{1}
      {,}{{{\color{punct}{,}}}}{1}
      {\{}{{{\color{delim}{\{}}}}{1}
      {\}}{{{\color{delim}{\}}}}}{1}
      {[}{{{\color{delim}{[}}}}{1}
      {]}{{{\color{delim}{]}}}}{1},
}

\makeatletter
\renewcommand{\section}{\@startsection{section}{1}{\z@}{1.0ex plus 1.0ex minus 0.5ex}%
	{0.5ex plus 1ex minus 0ex}{\normalfont\normalsize\centering\scshape}}%
\renewcommand{\subsection}{\@startsection{subsection}{2}{\z@}{1.0ex plus 1.0ex minus 0.5ex}%
	{0.5ex plus 1ex minus 0ex}{\normalfont\normalsize\itshape}}%
\makeatother

\begin{document}

\title{\LARGE \bf
Using The Concept Hierarchy for Household Action Recognition*
}

\author{Andrei Costinescu, Luis Figueredo and Darius Burschka$^{1}$
\thanks{*This work was supported by the Lighthouse Initiative Geriatronics by StMWi Bayern (Project X, grant no. 5140951).}
\thanks{$^{1}$All authors are with the School Of Computation, Information, and Technology at the Technical University of Munich. 
        {\tt\footnotesize \{andrei.costinescu, luis.figueredo, burschka\}@tum.de}}%
}

\markboth{IEEE Transactions on Knowledge and Data Engineering \commenting{,~Vol.~18, No.~9, September~2020}}%
{How to Use the IEEEtran \LaTeX \ Templates}

\maketitle

\begin{abstract}

We propose a method to systematically represent both the static and the dynamic components of environments, i.e. objects and agents, as well as the changes that are happening in the environment, i.e. the actions and skills performed by agents. Our approach, the Concept Hierarchy, provides the necessary information for autonomous systems to represent environment states, perform action modeling and recognition, and plan the execution of tasks. Additionally, the hierarchical structure supports generalization and knowledge transfer to environments. 
We rigorously define tasks, actions, skills, and affordances that enable human-understandable action and skill recognition.

\end{abstract}

\begin{IEEEkeywords}
knowledge modeling, 
and perceptual reasoning.
\end{IEEEkeywords}

\thispagestyle{empty}
\pagestyle{empty}

\section{Introduction}

Making sense of the world in which we, humans, live is a difficult problem. It is even more difficult for a robotic system. Acting intelligently and interacting with objects for purposeful changes in the environment requires both knowledge of objects, agents, actions, skills, and tasks as well as algorithms that leverage this knowledge and take into account the task, the abilities of the performing agent(s), and the environment in which the task is to be performed.

\commenting{
\toShorten{
From a very young age, we, humans, interact with the world and with the objects inside it. We start by observing the geometry of the world and discover that some parts of the environment can move when touched and thus we learn about object segmentation and recognition. Our toys as babies are specially designed for learning how to segment and recognize objects, and we start to remember individual shapes that later become objects. Every individual object is at first an instance, different than every other object by its shape. Through interaction with other objects or through the objects' physical properties such as color, taste, smell, and others, we start clustering objects into concepts. The set of concepts is ever-expanding and the concepts' individual definitions are ever-changing through various interactions within the world. 
}
}

\begin{figure}[t!]
    \centering
    \includegraphics[width=1\linewidth]{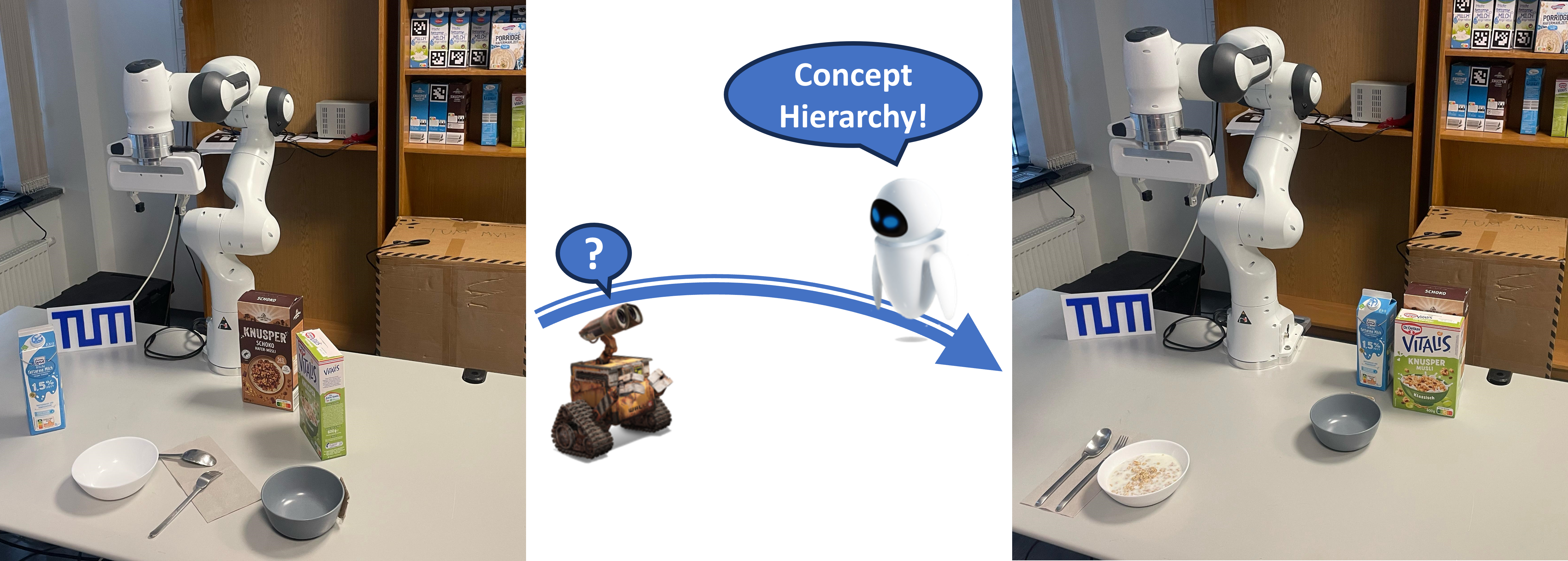}
    \caption{"How to transform the left environment into the right one?" The knowledge in the Concept Hierarchy enables household robots to represent environments and to create a plan to execute tasks.} \label{fig:concept_hierarchy_teaser}
\end{figure}

It is the goal of this paper to present a comprehensive framework for knowledge representation, including objects, agents, actions, skills, and tasks, that aids in the understanding of and participation in the environment dynamics, i.e. the environment changes and their causes. Thus, the framework can be used in, but is not limited to, household service robotics as in Figure \ref{fig:concept_hierarchy_teaser}. Applications of the framework include task planning, environment modeling, and action recognition.

Intelligent agents reason about perceived data in an environment and decide to act in the environment to fulfill a goal \cite{ai_book_1}. Structuring and representing knowledge for artificial agents is an important domain of artificial intelligence. One such technique is a \textbf{Semantic network}, also known as an ontology, which defines knowledge as concepts and semantic relations between them. 
OWL, the web ontology language, is a standard for writing ontologies \cite{w3c_owl}. 
In OWL, concepts, instances, properties, and relations between concepts can be defined. Properties have associated value types and one can define custom value types, such as vectors or matrices. However, inference with custom data types is not supported, and defining functions to modify the values of custom types is not easy. Composing functions to create new ones is not supported. 

Knowrob \cite{knowrob} extends the ontology in \cite{cyc_ontology} and aims to represent knowledge for executing robotic skills. 
Based on OWL, it is thus quite cumbersome to define functions in the knowledge representation itself that check if an action or a skill is active in the environment.
Furthermore, there is no clear distinction between a task, an action, and a skill.
Also, knowledge of verifying if skills are active is not represented in Knowrob; only knowledge about task execution.

The authors of \cite{rtpo} define a robot task planning ontology in which tasks are represented as linear action sequences. 
Their action definition as "atomic actions" is circular and resembles the definition of a motion primitive, implying that actions have different meanings for different robots. We clarify this in \ref{ssec:actions_def} and \ref{ssec:skills_def} by providing a concrete definition for actions.


The authors of \cite{light_nonmonotonic} represent knowledge as a conceptual hierarchy that can specify default values and exceptions, thus being a non-monotonic knowledge base. 
We also employ the Principle of Specificity described in \cite{sep-logic-nonmonotonic} to model default values and exceptions for concept properties, e.g. for skill properties. 
\commenting{
Finally, we overcome the limitations described in \cite{light_nonmonotonic} that 
programming languages such as Java and C++ are not suited for representing conceptual hierarchies because of their implicit closed-world assumption and the difficulty of updating the knowledge base.
}

\commenting{
\begin{itemize}
    \item \href{https://www.researchgate.net/publication/328249457_A_Review_of_Knowledge_Bases_for_Service_Robots_in_Household_Environments}{A Review of Knowledge Bases for Service Robots in Household Environments};
    \item Karinne Ramirez Amaro 
\end{itemize}
}

Probabilistic learning methods have become popular in various applications, including action recognition \cite{human_action_recognition_survery, hao_pgcn}. Their advantage is a fast generation of possible action hypotheses and a human-like description or segmentation of the recognized actions in a demonstration. 
Such approaches lack, however, an understanding of \textbf{why} motions "look like" actions.

A model-based approach enables richer uses of knowledge, such as recognizing failures and pinpointing the reason for an unsuccessful skill execution thanks to a model of the physical world. Furthermore, a model-based approach can help identify the missing steps or needed circumstances to successfully execute a skill. Having a model of an action enables verification of whether it is truly happening in the scene and not just if it "looks like" the action is executed. Finally, compared to neural networks, a model-based approach \textbf{can} model the states of the environment and make long-term temporal decisions based on the properties of objects present in the environment, so we have decided not to use a learning approach for action recognition.


\subsection{Contributions}
Our new knowledge representation allows us to represent both the "things" in an environment, i.e., objects and agents, and the changes happening in the environment.
We provide structured definitions of the represented knowledge and fill the Concept Hierarchy (\textbf{CH}) with concepts and instances that can be used to execute skills on a robotic agent and to recognize and understand the changes in the environment. 
We rigorously define the difference between tasks, actions, and skills (terms that usually do not have a clear meaning or differentiation in robotics literature) and show how affordances are represented in the \textbf{CH}.
Our implementation is written in C++ and supports dynamic type changes of instances, i.e., instances adding or removing concepts at runtime. It also enables data updating and programmatic restructuring of the existing knowledge.

\subsection{Structure}
Section \ref{sec:approach} presents our modeling of concepts, instances, objects, agents, actions, skills, tasks, and contexts in the \textbf{CH}. The \textbf{CH}'s use for action and skill recognition is described in Section \ref{sec:applications}, followed by the conclusion and future work.

\section{The Content of the Concept Hierarchy}\label{sec:approach}
\begin{figure}[t!]
    \centering
    \includegraphics[width=1\linewidth]{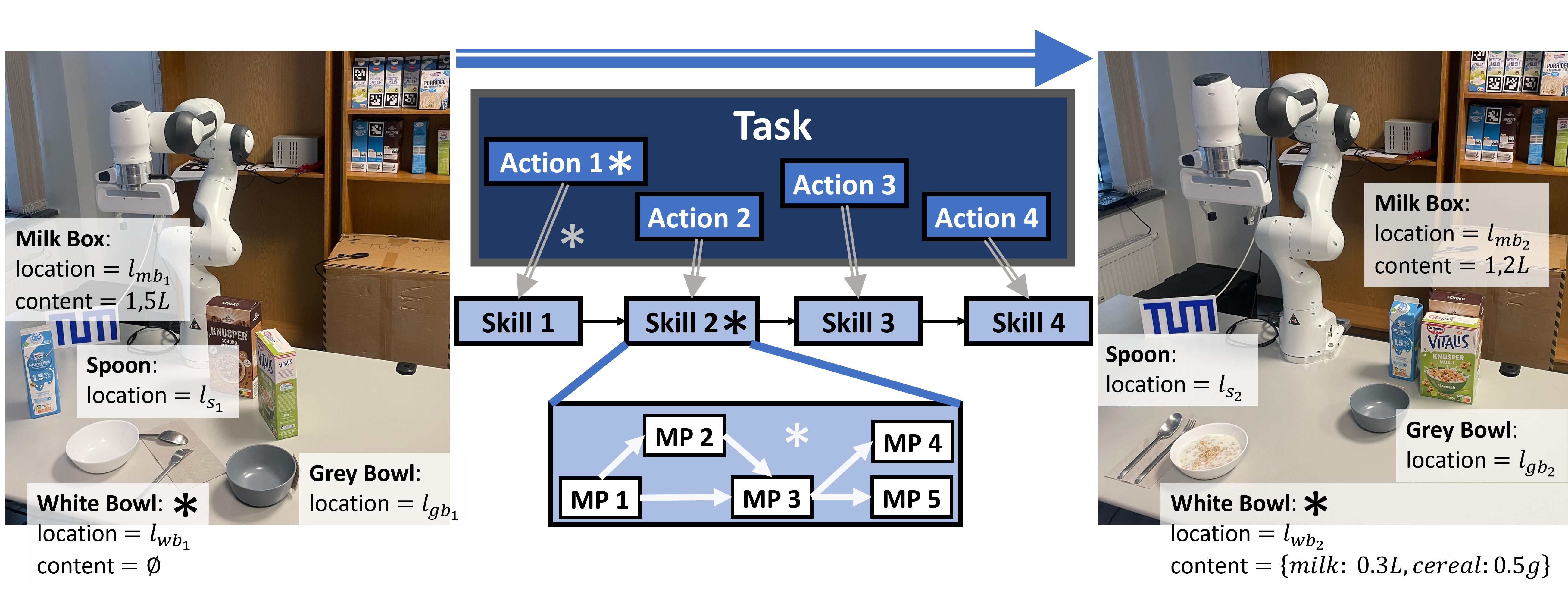}
    \caption{The task of transforming the left environment to the desired right one is divided into actions that are correlated to skills, which are composed of Motion Primitives (MP). \textbf{*} marks knowledge from the \textbf{CH}: action and skill definitions, skill-to-action association, sequencing of motion primitives for skill execution, and entity properties.} \label{fig:objects_and_tasks_domain}
\end{figure}

The Concept Hierarchy (\textbf{CH}) is a knowledge modeling framework to represent the information needed for a particular application or within an application domain. Figure \ref{fig:objects_and_tasks_domain} shows an application domain where the \textbf{CH}'s data is useful.

To turn the left environment into the right one, one must first describe the state of both environments. Then, one determines the differences between the two environments, which lastly results in a sequence of changes for an agent to execute.

For describing the environment state, it is not enough to just segment the environment's geometry into clusters. The clusters must have semantic information attached to them and properties that differentiate the clusters from others of similar type or shape. We thus introduce \textit{Concepts} that define the properties of different cluster types, i.e. \textit{Container}, \textit{MilkBox}, \textit{Bowl}, \textit{CerealBox}, \textit{Spoon}, \textit{Human}, \textit{Robot}, etc. The actual properties values are not set by \textit{Concepts}, but by \textit{Instances}, the knowledge containers. The environment state is the collection of all properties of entities inside it, i.e. \textit{Objects} and \textit{Agents}. 

The differences between the entity properties in the start and end environment are formalized and represented as \textit{Actions} in the \textbf{CH}. One \textit{Action} represents one change in an entity's property. In Figure \ref{fig:objects_and_tasks_domain}, the \underline{location} property of the \textit{Milk Box} is different. This is represented by the \textit{ChangeLocation} \textit{Action}. For performing this change (physically) in the environment, we represent \textit{Skills} in the hierarchy, which are agent-dependent. For executing the \textit{ChangeLocation} in the environment, a \textit{Human} can choose from a large number of \textit{Skills}, such as \textit{Transporting}, \textit{Throwing}, \textit{Rolling}, \textit{Pushing}, etc. \textit{Actions} and \textit{Skills} are also \textit{Concepts} and have parameters that describe the change and the way of performing it in the environment.

\begin{figure}[t!]
    \centering
    \includegraphics[width=1\linewidth]{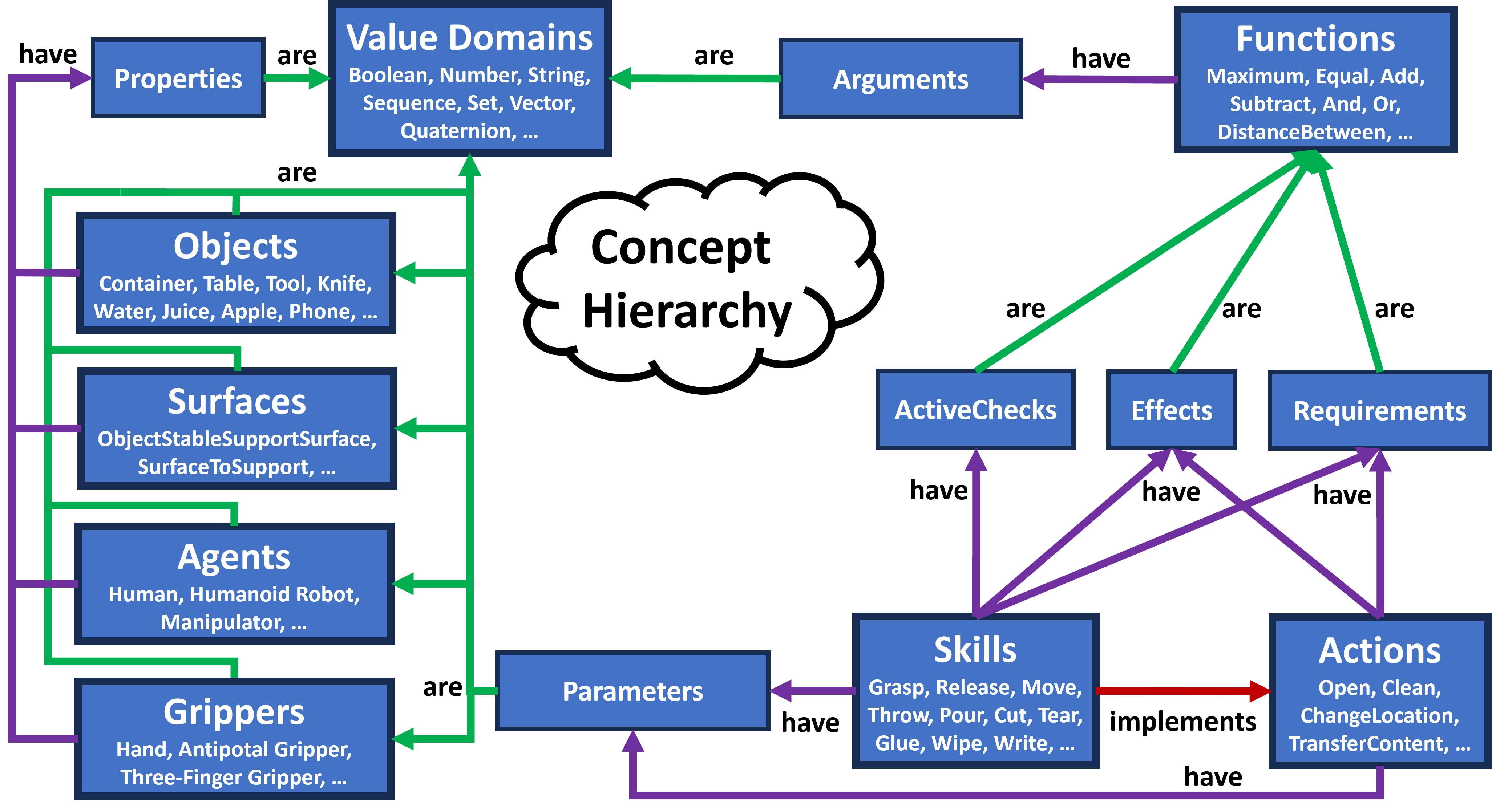}
    \caption{The \textbf{CH} contains \textit{Objects}, \textit{Surfaces}, \textit{Agents}, and \textit{Grippers}. Their specific properties are modeled as \textit{ValueDomains}. \textit{Skills} and \textit{Actions} contain \textit{ValueDomain} parameters, including \textit{Objects} and \textit{Agents}. \textit{Skills} perform \textit{Actions} in an environment. They have requirements from and effects on their parameters and can check whether a skill is active. These are modeled as \textit{Functions} with \textit{ValueDomain} arguments.} \label{fig:concept_hierarchy_components}
\end{figure}

After determining the necessary changes, a \textit{Task} planner sequences, prioritizes and optimizes their execution in the environment. This final \textit{Task} execution plan is then distributed to the agent(s) that will transform the environment into its desired state. The next sections present how the stored knowledge is grouped and structured inside the Concept Hierarchy. Figure \ref{fig:concept_hierarchy_components} presents the interaction of the knowledge in the \textbf{CH}.

\subsection{\textbf{Concepts}, \textbf{Instances} and \textbf{ValueDomains}: Knowledge Definitions, Containers, and Data Types} \label{ssec:def_concept_instance}
In the example of Figure \ref{fig:objects_and_tasks_domain}, the environment consists of many \textbf{Instances}, some of them being the spoon, the white and grey bowls, and the milk box. Respectively, the entities are instances of the \textit{KitchenUtensil}, \textit{Bowl}, and \textit{MilkContainer} concepts; the latter two are subconcepts of \textit{Container}.

Some differences between the left and right of Figure \ref{fig:objects_and_tasks_domain} are that the milk box has less content and a different location, the grey bowl has a different location, and the white bowl has now milk and cereal as contents. The terms \underline{content} and \underline{location} are instance properties. They are defined in the \textit{Container} and \textit{PhysicalEntity} concepts respectively. A property also defines the range of values that it can have. This value range is modeled as \textit{ValueDomains}, knowledge data types.

The definition of a \textit{Concept} in the \textbf{CH} includes which properties pertain to the concept and their associated \textit{ValueDomain}. E.g., the \underline{location} property of a \textit{PhysicalEntity} has a \textit{Location} \textit{ValueDomain}. A \textit{Location} is a pose displacement relative to a coordinate frame. And the \textit{ValueDomain} of the \underline{content} property of a \textit{Container} is a \textit{List} of \textit{PhysicalEntities}. 

\textit{Concepts} are organized hierarchically and inherit the property definitions of their parent \textit{Concepts}. The \textit{Concept} itself does not specify a value for its properties; this is done by \textit{Instances}, the containers of knowledge. Unspecified property values are set to the \textit{UNKNOWN} value, making the world model follow an open-world assumption. \textit{ValueDomains} are not bound to real data types used in current programming languages, to not tie the \textbf{CH} to a specific programming language, and to let the domain expert implement the optimal data structures for the particular application of need.

\subsection{\textbf{Functions}: Mathematical Modifiers of \textit{ValueDomains}}\label{ssec:functions}
The changes in the property values of concept properties are carried out by \textit{Functions}. \textit{Functions} have arguments \textit{ValueDomains} and represent operations that can be executed on them. E.g., \textit{Addition}, \textit{Multiplication} of two \textit{Numbers}, \textit{Incrementing} one \textit{Number}, \textit{Inserting} one item to a \textit{List}, and so on. The meaning of the \textit{Function}, i.e., what it actually does with its inputs, is, as with \textit{ValueDomains}, not necessarily specified in its definition of the \textbf{CH}. The application designer's job is to implement the meaning of the \textit{Function} in a desired programming language using the data types selected as \textit{ValueDomains}. The \textit{Function}'s definition inside the Concept Hierarchy only needs to specify the function's interface and its name; similar to what a header file in C++ defines.

There are \textit{Concept} properties that form a dependency cycle: \textit{Agents} have \textit{Grippers}, and a \textit{Gripper} belongs to an \textit{Agent}. Sometimes, changing an instance property must have an effect on a different property in a related instance, e.g., removing a gripper from an agent must update the \underline{belongingAgent} of the removed gripper. To model this, \textit{Concepts} define \textit{hooks} on properties, which are represented as \textit{Functions}.

\subsection{\textbf{Actions}: Modifiers of Knowledge Containers}\label{ssec:actions_def}
\textit{Actions} are used to formally model the changes in Figure \ref{fig:objects_and_tasks_domain}. We define an \textit{Action} as one change in the properties of an instance. E.g., the \textit{ChangeLocation} \textit{Action} changed the \underline{location} property of the milk box, and \textit{TransferContent} took 0.3L of milk from the milk box into the white bowl.

\textit{Actions} do not describe how the change was achieved in the environment; they model \textbf{what} the change was. Actions have preconditions from and effects on their entity parameters (i.e. the instances on which the action is to be performed), which are modeled as \textit{Functions}.

\subsection{\textbf{Skills}: Action Implementation in Environments}\label{ssec:skills_def}
The \textit{Concept} describing \textbf{how} the change represented by an \textit{Action} was performed is a \textit{Skill}. E.g., \textit{Pushing}, \textit{Pulling}, or \textit{Carrying} are \textit{Skills} that could have been used to \textit{ChangeLocation} of the milk box and \textit{Pouring} or \textit{Scooping} could have been used to \textit{TransferContents}. \textit{Skills} are agent-, instance-, and environment-specific. They depend on the capabilities of the executing agent, the instance's properties, and the environment's configuration (e.g. obstacles) for successful execution. In addition to the preconditions and effects, \textit{Skills} define two more \textit{Functions}: the check for the skill being active and for it being successfully executed in the environment.

Being particular to certain agents, instances, and/or environments, a \textit{Skill}'s property \textit{ValueDomain} definition can include concepts that instances must not be sub-concepts of. \textit{Flying}, a \textit{Skill} associated with the \textit{ChangeLocation} \textit{Action}, has \textit{Birds} as the \textit{ValueDomain} for its entity property and also the restriction of the entity not being a \textit{Penguin}, known not to be able to fly.

The \textit{Skill} concept defines the \textit{manipulations} property: a triple of agent-gripper-object entities (an agent uses a gripper to manipulate an object). \textit{Skill} subconcepts define a default value for the \textit{manipulations} property, that all \textit{Skill} instances will have unless overwritten by the instance. E.g., \textit{Pouring}, the skill describing the process of tilting the \textit{Container} $from$ located in the \textit{Agent} $a$'s \textit{Gripper} $g$ over the \textit{Container} $into$, defines the \underline{manipulation}'s default value to be $\{\left(a, g, from\right)\}$. The subconcept of \textit{Pouring}, \textit{PouringWith2Grippers}, which has the $into$ \textit{Container} in the \textit{Agent} $a$'s second \textit{Gripper} $g_2$, overwrites \textit{Pouring}'s default value with $\{\left(a, g, from\right), \left(a, g_2, into\right)\}$. This feature of the \textbf{CH} makes it a non-monotonic knowledge base.

\textit{Skills} also have an \underline{actionAssociation} property, that creates the association between a \textit{Skill} and the, possibly multiple, \textit{Actions} that the \textit{Skill} implements in the environment.

\subsection{\textbf{Affordances}}\label{ssec:affordances} 
Determining or estimating entity affordances is still an open problem in robotics. Affordances indicate which actions and skills can be performed on the entity. I.e., into which action or skill property can the entity be substituted. We represent the action-affordances of an entity $e$ as the set of pairs $Aff\left(e\right) = \{ \left(a, p \right) |\ a \in Actions, p \in E\left(a\right),$ $e$ is a subconcept of the property $p$ of the action $a\}$, where $E\left(a\right)$ is the tuple of $a$'s entity-properties. Skill-affordances are defined similarly. 

Affordances are thus related to the definition of action and skill properties. The \textit{Concepts} that are the \textit{ValueDomains} of action and skill properties thus know as which properties their instances can be used. Therefore, affordances do not need to be stored explicitly in the \textbf{CH}; they are the entity-\textit{ValueDomains} of the action and skill properties.

\subsection{\textbf{Motion Primitives}, \textbf{Abilities}: Building Blocks of Skills}\label{ssec:motion_primitives}
Abilities are the building blocks of skills. They are the generalization of motion primitives; a motion is not always needed as a part of skill execution. For \textit{Walking}, the agent must \underline{perceive} the environment to determine a collision-free walking path. Perception does not always require a motion from the agent. Another ability is \underline{waiting} until something has happened, which does not involve any motion. Yet, it is an important part of a skill's execution, e.g. during collaboration with multiple agents. \textit{Skill} execution is thus converting its parameter values into correct parameters for the executing agent's abilities.

\subsection{\textbf{Tasks}: Multiple Changes in the Environment}\label{ssec:task_def}
In our work, a task is a desired state of the environment. A state of the environment specifies instance properties. This desired state, when compared to the current state of the environment, results in differences in entity properties: a set of instance property changes, i.e. a set of actions. The task can also define constraints on the ordering of actions or constraints on their execution that must be considered by a planner when parsing the set of actions into a sequence of skills.

In Figure \ref{fig:objects_and_tasks_domain}, the task is to transform the environment into the state on the right. By comparing the two environments, the \textbf{task} planner determines the differences in the values of instance properties in the environment and determines the \textit{Actions} that have been done. It is then a smart agent's job to associate the \textit{Actions} with \textit{Skills} that implement the \textit{Action}, that fit the abilities of the agent, and that are executable in the environment configuration.

The following section presents an example of using the \textbf{CH}'s knowledge in a household setting.

\begin{figure*}[t!]
    \centering
    \includegraphics[width=1.\linewidth]{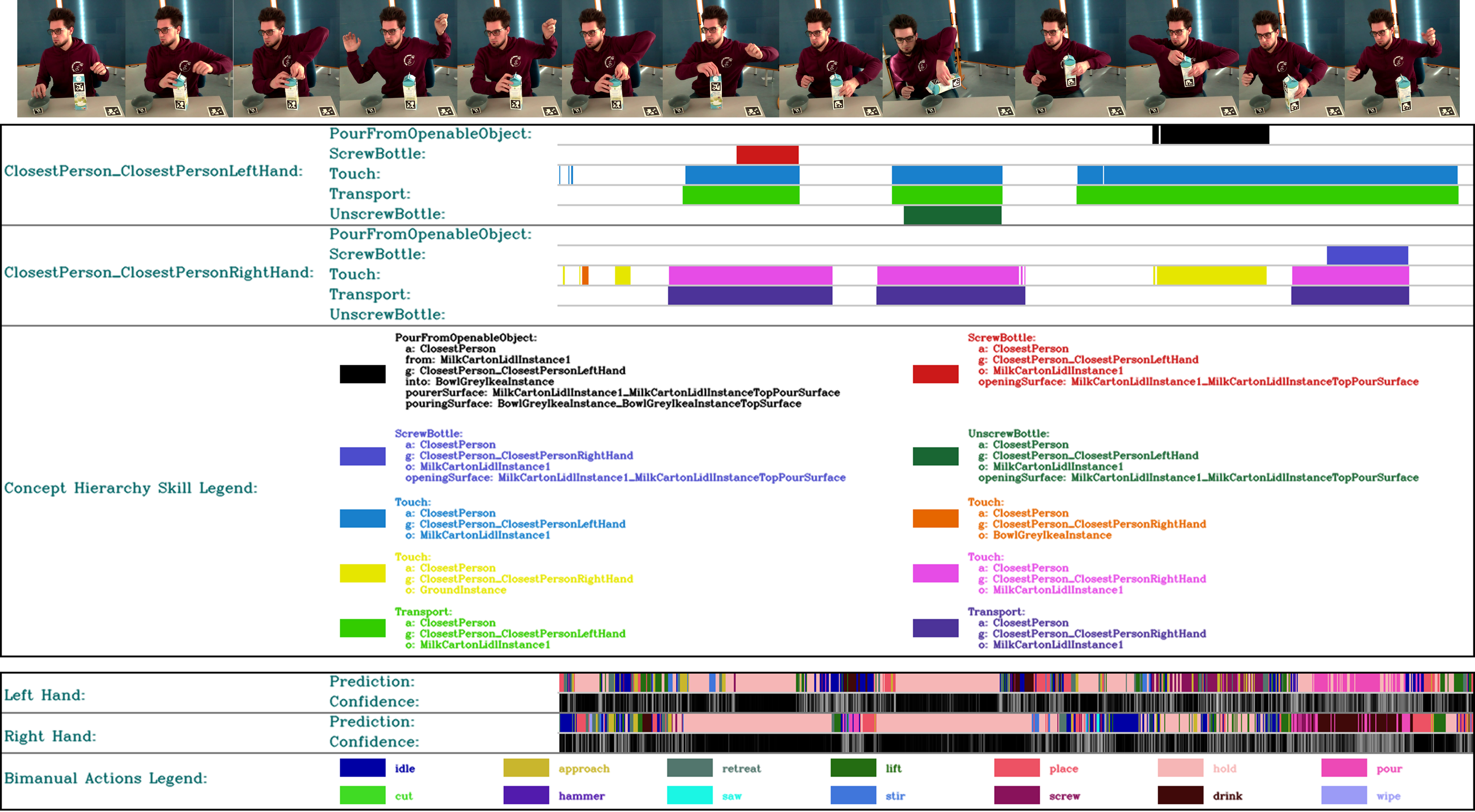}
    \caption{Skill recognition result of a bimanual task of pouring milk into a bowl. The demonstration is composed of closing and then opening a milk box with the left hand, pouring milk into the bowl, and closing the milk box with the right hand. The upper figure presents the \textit{Skill} instances our method recognizes for each hand and \textit{Skill} type. The colors help distinguish \textit{Skills} of the same type with different parameters. The lower figure shows the results of \cite{hao_pgcn} that was trained on the Bimanual Actions dataset \cite{bimacs_dataset} (darker confidence = higher softmax output).} \label{fig:action_skill_recognition}
\end{figure*}

\section{Action and Skill Recognition}\label{sec:applications}
\begin{figure}[t!]
    \centering
    \includegraphics[width=1\linewidth]{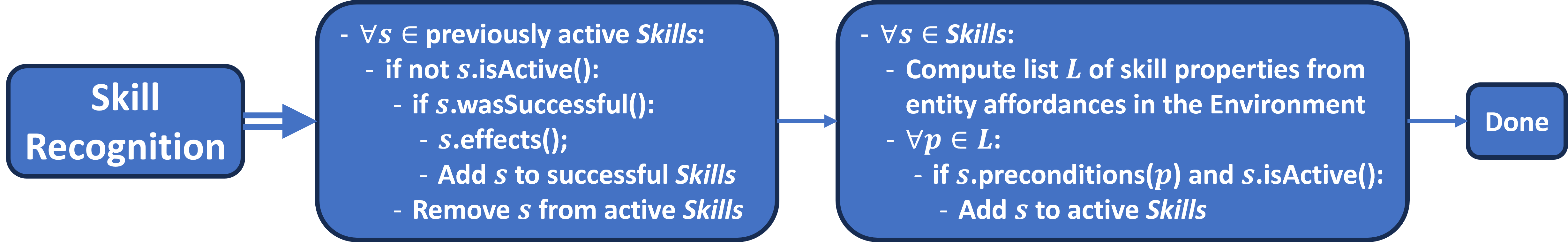}
    \caption{The procedure for \textit{Skill} recognition.} \label{fig:action_skill_recognition_process}
\end{figure}

We present the \textbf{CH}'s use in recognizing actions and skills in a household environment. We have used OpenPose \cite{openpose} and AprilTags \cite{aprilTag} to get human hand 3D positions and object 3D poses from a Realsense \cite{realsense} camera. 
Having a system based on visual input, the functions checking if a \textit{Skill} is active are visual, geometrical relations between object and agent instances. With each new camera frame, the objects and agents are recognized and localized inside the environment. Then, the skill recognition process starts and is presented in Figure \ref{fig:action_skill_recognition_process}.
Using the \underline{actionAssociations} property, the successfully executed skills determine the actions in the environment.

Figure \ref{fig:action_skill_recognition} compares the observed skills from our method with \cite{hao_pgcn} on a milk pouring example where the milk box is closed and then opened with the left hand and closed again with the right hand. The \textit{Transport} skill defines its $o$ parameter as a \textit{MovableObject}, which is why the \textit{MilkCartonLidlInstance} was transported and the \textit{GroundInstance} not, even if touched with the right hand.

Compared to a learning method that outputs probabilities of recognized skills, our method \underline{certainly} determines successfully executed skills. It can also determine multiple skills active at the same time as well as with which objects the agent interacted. The probabilistic method has misdetections possibly explained by the agent executing similar motions with his hand that "look like" other actions, even if no object was touched. Even if the agent moves his hand in a lifting manner, which the learning-based method recognizes, it fails to understand that no object is lifted or even held in the hand. This is the advantage of a model-based method: the verifiability and explainability of a decision; lifting did not occur because the hand did not touch an object in that frame.
\section{Conclusion}

This paper describes the building blocks of our knowledge modeling framework for household applications, the Concept Hierarchy. This definition of concepts and instances efficiently represents household objects in manipulation tasks. It supports dynamic changes of concept properties due to performed actions and enables inheritance and overwriting of default property values. Our parametrizable skill and action definition allows both a general and restrictive specification of instances, e.g. the \textit{Flying} \textit{Skill}'s agents are \textit{Birds} except \textit{Penguins}.


Future work includes using the defined knowledge for learning and executing \textit{Tasks} and creating a planner to leverage the intrinsic liberties of \textit{Task} descriptions. Furthermore, we plan to infer unknown instance properties from the open-world assumption by observing the effects and preconditions of performed actions on the instance.

\bibliographystyle{IEEEtran}
\bibliography{references}

\end{document}